\def\year{2020}\relax
\documentclass[letterpaper]{article} 
\usepackage{aaai20}  
\usepackage{times}  
\usepackage{helvet} 
\usepackage{courier}  
\usepackage[hyphens]{url}  
\usepackage{graphicx} 
\usepackage{graphicx}  
\usepackage{pgfplots, pgfplotstable,booktabs}
\usetikzlibrary{positioning}

\definecolor{nfs}{RGB}{37,165,203}
\definecolor{ufs}{RGB}{253,177,26}
\definecolor{cfs}{RGB}{175,32,67}
\definecolor{c1}{RGB}{102,0,102}
\definecolor{c2}{RGB}{0,153,153}

\title{A Benchmark Dataset of Check-worthy Factual Claims}
\author{Fatma Arslan,\textsuperscript{\rm 1} Naeemul Hassan,\textsuperscript{\rm 2} Chengkai Li,\textsuperscript{\rm 1} Mark Tremayne\textsuperscript{\rm 3}\\
\textsuperscript{\rm 1}Department of Computer Science and Engineering, University of Texas at Arlington\\
\textsuperscript{\rm 2}College of Information Studies, University of Maryland\\
\textsuperscript{\rm 3}Department of Communication, University of Texas at Arlington\\
fatma.dogan@mavs.uta.edu, nhassan@umd.edu, cli@uta.edu, tremayne@uta.edu}

\begin{document}

\maketitle

\begin{abstract}
 In this paper we present the ClaimBuster dataset of $23,533$ statements extracted from all U.S. general election presidential debates and annotated by human coders.  The ClaimBuster dataset can be leveraged in building computational methods to identify claims that are worth fact-checking from the myriad of sources of digital or traditional media. The ClaimBuster dataset is publicly available to the research community, and it can be found at \url{http://doi.org/10.5281/zenodo.3609356}.
\end{abstract}

\section{Introduction}
\label{sec-introduction}

Misinformation is a growing problem around the world~\cite{bradshaw2019global}. Journalists and fact-checkers work constantly to identify and correct misinformation and to communicate their work as soon as possible. However, given the amount of information being created daily and the  limited resources available to  journalists, it has become almost impossible keep up this critical work. Researchers from various disciplines~\cite{Babakar2016,relevantdocumentdiscovery,popupfactckecking-compjour19}, particularly computer science~\cite{hassan2017claimbuster,Miranda2019,jo2019verifying}, have come forward to create automated fact-checking tools. One of the key elements in the fact-checking process is automatically assessing the check-worthiness of a piece of information. Such an assessment can not only assists the journalists with providing them the most check-worthy claims from an interview or debate but also lessens the potential of human bias in claim selection. However, to have an accurate automated check-worthiness assessment, it is imperative to have a carefully annotated ground-truth dataset that can fuel a machine learning algorithm to predict the check-worthiness of a statement. 

In this paper, we present a dataset of claims from all U.S. presidential debates (1960 to 2016) along with human-annotated check-worthiness label. It contains $23,533$ sentences where each sentence is categorized into one of three categories- non-factual statement, unimportant factual statement, and check-worthy factual statement. These sentences have been labeled by 101 coders over a 26 months period in multiple phases.

This dataset has been used to develop the first-ever end-to-end automated fact-checking system, ClaimBuster~\cite{hassan2017toward,hassan2017claimbuster}. It has been used to study how an automated check-worthiness detector fares compared to human judgements~\cite{hassan2016comparing}. Also, it has been used to deliver check-worthy factual claims filtered from a variety sources including PolitiFact,~\footnote{\url{https://www.politifact.com/}} one of the leading fact-checking organization in the United States~\cite{humantouch-compjour19}. Through this paper, we make the dataset publicly available.

In the following sections, we describe the preparation process of the dataset, present descriptive statistics of the dataset, suggest possible use cases, and explain different fairness policies we have followed while developing this dataset.

\section{Related Works}
\label{sec-relatedwork}

\begin{figure*}[t]
\centering
  \begin{tikzpicture}
    \centering
    \begin{axis}[
        width  = 0.7*\textwidth,
        height  = 0.2*\textwidth,
        ylabel= {\textbf{Number of Sentences}},
        label style={font=\small},
        enlargelimits=0.05,
        ybar=2*\pgflinewidth,
        bar width=.15cm,
        x tick label style={font=\small, rotate=90},
        y tick label style={font=\small},
        symbolic x coords={1960-09-26, 1960-10-07, 1960-10-13, 1960-10-21, 1976-09-23, 1976-10-06, 1976-10-22, 1980-09-21, 1980-10-28, 1984-10-07, 1984-10-21, 1988-09-25, 1988-10-13, 1992-10-11, 1992-10-15, 1992-10-19, 1996-10-06, 1996-10-16, 2000-10-03, 2000-10-11, 2000-10-17, 2004-09-30, 2004-10-08, 2004-10-13, 2008-09-26, 2008-10-07, 2008-10-15, 2012-10-03, 2012-10-16, 2012-10-22, 2016-09-26, 2016-10-09, 2016-10-19 }, 
        xtick align=inside,
        ymin=100,
        ymax=1000,
        xtick=data]
    \addplot [fill=c1]
        coordinates {(1960-09-26,425)(1960-10-07,432)
        (1960-10-13,372)(1960-10-21,456)(1976-09-23,483)(1976-10-06,472)(1976-10-22,471)(1980-09-21,337)(1980-10-28,469)(1984-10-07,599)(1984-10-21,528)(1988-09-25,721)(1988-10-13,674)(1992-10-11,823)(1992-10-15,863)(1992-10-19,849)(1996-10-06,931)(1996-10-16,874)(2000-10-03,885)(2000-10-11,822)(2000-10-17,785)(2004-09-30,873)(2004-10-08,942)(2004-10-13,830)(2008-09-26,803)(2008-10-07,714)(2008-10-15,737)(2012-10-03,837)(2012-10-16,923)(2012-10-22,858)(2016-09-26,1012)(2016-10-09,848)(2016-10-19,885)};
    \end{axis}
    \end{tikzpicture}
  \caption{Sentence distribution among presidential debates} 
  \label{fig:sentence_dist}

\centering
  \begin{tikzpicture}
    \centering
    \begin{axis}[
        width  = 0.7*\textwidth,
        height  = 0.2*\textwidth,
        ylabel= {\textbf{Avg Sentence Length}},
        label style={font=\small},
        enlargelimits=0.05,
        ybar=2*\pgflinewidth,
        bar width=.15cm,
        x tick label style={font=\small, rotate=90},
        y tick label style={font=\small},
        symbolic x coords={1960-09-26, 1960-10-07, 1960-10-13, 1960-10-21, 1976-09-23, 1976-10-06, 1976-10-22, 1980-09-21, 1980-10-28, 1984-10-07, 1984-10-21, 1988-09-25, 1988-10-13, 1992-10-11, 1992-10-15, 1992-10-19, 1996-10-06, 1996-10-16, 2000-10-03, 2000-10-11, 2000-10-17, 2004-09-30, 2004-10-08, 2004-10-13, 2008-09-26, 2008-10-07, 2008-10-15, 2012-10-03, 2012-10-16, 2012-10-22, 2016-09-26, 2016-10-09, 2016-10-19 },
        xtick align=inside,
        ymin=5,
        ymax=25,
        xtick=data]
    \addplot [fill=c2]
        coordinates {(1960-09-26,22.1812)(1960-10-07,22.5810)(1960-10-13,23.4140)(1960-10-21,20.8816)(1976-09-23,24.4389)(1976-10-06,22.4386)(1976-10-22,23.7665)(1980-09-21,24.7685)(1980-10-28,25.6311)(1984-10-07,19.6127)(1984-10-21,20.7898)(1988-09-25,18.0014)(1988-10-13,18.6261)(1992-10-11,17.2321)(1992-10-15,16.8088)(1992-10-19,17.5053)(1996-10-06,16.6122)(1996-10-16,16.1167)(2000-10-03,16.3017)(2000-10-11,16.7932)(2000-10-17,16.1669)(2004-09-30,15.0389)(2004-10-08,14.6051)(2004-10-13,15.5253)(2008-09-26,18.2279)(2008-10-07,18.8669)(2008-10-15,18.6621)(2012-10-03,17.8411)(2012-10-16,16.5049)(2012-10-22,17.8834)(2016-09-26,14.1354)(2016-10-09,15.1333)(2016-10-19,14.4045)};
    \end{axis}
    \end{tikzpicture}
  \caption{Average sentence length in words per debate} 
  \label{fig:sentence_length}
\end{figure*}

Researchers have attempted to prepare datasets of check-worthy factual claims to assist automated fact-checking. For instance, Nakov et al.~\cite{clef2018checkthat:overall} developed a dataset of check-worthy factual claims from the 2016 U.S. presidential debate. To determine the check-worthiness of statements, the authors used available fact-checks of the debate by a fact-checking organization, FactCheck.org. If FactCheck.org has checked a statement from the debate, the dataset labels that statement as check worthy; otherwise not. While this strategy ensures that their check-worthy statements are indeed picked by professional fact-checkers it does not resolve the question of whether selection bias of a single organization may have tainted the quality of the dataset. Our strategy for annotation considers input from multiple high-quality, trained coders. This decreases the chance of having a dataset with a bias towards certain ideology. Also, unlike the dataset of ~\cite{clef2018checkthat:overall}, that had 2016 debates and several political speeches of that time, we annotated all the U.S. general election presidential debates since 1960.

Patwari et al.~\cite{patwari2017tathya} prepared another dataset of check-worthy factual claims by combining the fact-checks of 15 2016 U.S. election primary debates from 9 fact-checking organizations (e.g., Fox News, NPR, CNN). Although having inputs from a range of fact-checking organizations reduces the chance of having a biased sample the dataset becomes specific to certain issues that were relevant during the 2016 presidential election. As our dataset covers a longer time-period, over 50 years, it captures more general issues and patterns that are relevant for assessing the check-worthiness of a broader array of claims.

\section{Transcript Extraction and Processing}
\label{sec-collection}

Candidate sentences were extracted from U.S. presidential debate transcripts.~\footnote{\url{https://www.debates.org/voter-education/debate-transcripts/}} The first general election presidential debate was held in 1960. Since then, there were a total of 15 presidential elections from 1960 to 2016. In 1964, 1968, and 1972, no presidential debate was held. There were 2 to 4 debate episodes in each of the remaining 12 elections. A total of 33 debate episodes spanned from 1960 to 2016. There are $32,072$ sentences spoken in these debates. We applied the following steps to prepare the candidate sentences to be labeled.

\begin{enumerate}
    \item Using parsing rules and human annotation, the speaker of the each sentence was identified. $26,322$ sentences are spoken by the presidential candidates, $4,292$ by the debate moderators, and $1,319$ by the questioners. There are $139$ sentences without a speaker name which were voice-over announcers at the start of the debate (i.e., ``September 26, 2008.'', ``The First McCain-Obama Presidential Debate'').  
    \item We only focused on the sentences spoken by the presidential candidates. Therefore, sentences spoken by the debate moderators, the questioners, and the announcers were discarded from further labeling.
    \item Another processing step was performed to filter very short sentences. We removed sentences shorter than 5 words. In total, $2,789$ sentences were discarded, which represent 8.69\% of the original dataset. 
\end{enumerate}

The resulting dataset (henceforth referred to as the \emph{ClaimBuster} dataset) contains $23,533$ labeled sentences. Figure~\ref{fig:sentence_dist} shows the distribution of the sentences among 33 debate episodes and Figure~\ref{fig:sentence_length} depicts the average length of sentences per debate. These figures show that although the number of spoken sentences increased in recent debates, they got shorter comparing to earlier debates.

\section{Annotation Procedure}
\label{sec-annotation}

\begin{figure*}[t]
\centering
\includegraphics[width=0.9\textwidth]{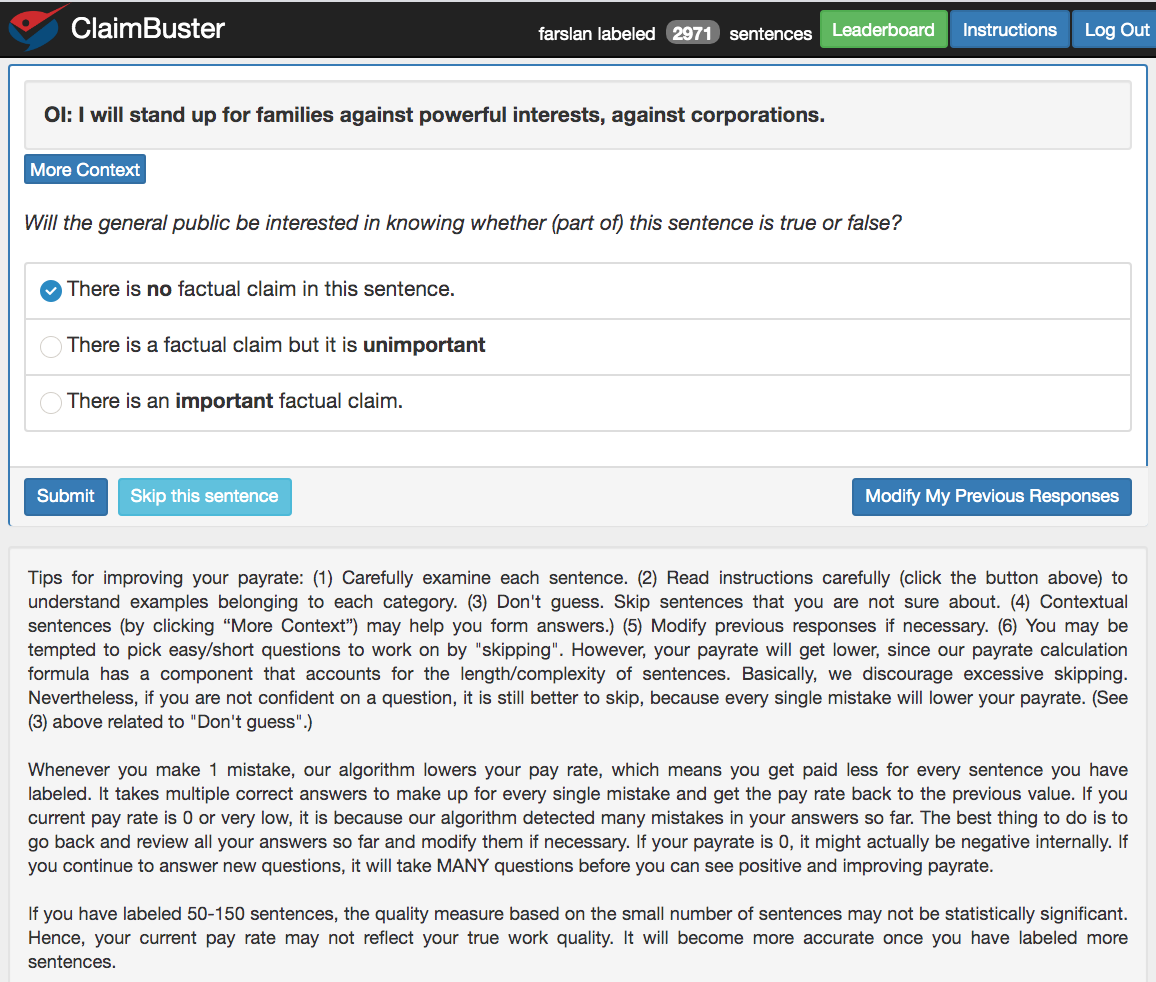}
\caption{Data collection interface}
\label{fig:data_collection}
\end{figure*}

\subsection{Annotation Guideline}
We categorize the sentences from the \emph{ClaimBuster} dataset into three groups. Below, we define each category, along with examples.

\noindent \textbf{Check-worthy Factual Sentence (CFS): }These sentences contain factual claims that the general public will be interested in learning about their veracity.  Journalists look for these types of claims for fact-checking. Some examples are:
\begin{itemize}
    \item In the last month, we've had a net loss of one hundred and sixty-three thousand jobs.
    \item We've spent \$4.7 billion a year in the State of Texas for uninsured people.
    \item When they tried to reduce taxes, he voted against that 127 times.
    \item China and India are graduating more graduates in technology and science than we are.
    \item My opponent opposed the missile defenses.
\end{itemize}

\noindent \textbf{Unimportant Factual Sentence (UFS): }These are factual claims but not check-worthy. In other words, the general public will not be interested in knowing whether these sentences are true or false. Fact-checkers do not find these sentences as significant for checking. A few examples are as follows:

\begin{itemize}
    \item I am a son of a Methodist minister.
    \item Just yesterday, I was in Toledo shaking some hands in a line.
    \item Well, the Vice President and I came to the Congress together 1946; we both served in the Labor Committee.
    \item And I've got two daughters and I want to make sure that they have the same opportunities that anybody's sons have.
\end{itemize}

\noindent \textbf{Non-factual Sentence (NFS): }These sentences do not contain any factual claims. Subjective sentences (opinions, beliefs, declarations) and many questions fall under this category. Below are some examples.

\begin{itemize}
    \item The worst thing we could do in this economic climate is to raise people's taxes.
    \item I think the Head Start program is a great program.
    \item We need to cut the business tax rate in America.
    \item I'll get America and North America energy independent.
\end{itemize}

\begin{table*}[!ht]
\centering
\begin{tabular}{lcl}
\toprule
\multicolumn{1}{c}{\textbf{Sentence}} & \textbf{Label} & \multicolumn{1}{c}{\textbf{Explanation}} \\ \toprule
\begin{tabular}[c]{@{}l@{}}Well, you know, nailing down Senator Obama's vari-\\ous tax proposals is like nailing Jell-O to the wall.\end{tabular} & NFS & \begin{tabular}[c]{@{}l@{}}This statement does not contain any factual information. \\ It is a rhetorical expression.\end{tabular} \\ \midrule
I'm simply not going to do that. & NFS & \begin{tabular}[c]{@{}l@{}}This statement does not contain any factual information. \\ The speaker is making a promise and/or talking about \\ his/her future plan.\end{tabular} \\ \midrule
\begin{tabular}[c]{@{}l@{}}In addition to that, we've suffered because we haven't\\ had leadership in this administration.\end{tabular} & NFS & \begin{tabular}[c]{@{}l@{}}This statement does not contain any factual information. \\ It is about the speaker's opinion or position on a certain \\ topic.\end{tabular} \\ \midrule
I was Governor of Georgia for four years. & UFS & \begin{tabular}[c]{@{}l@{}}This statement contains factual information. However, \\ the general public would not be interested in checking \\ the presented factual claim.\end{tabular} \\ \midrule
\begin{tabular}[c]{@{}l@{}}In Puerto Rico this year, I met with six of the leading\\ industrial nations' heads of state to meet the problem\\ of inflation so we would be able to solve it before it\\ got out of hand.\end{tabular} & UFS & \begin{tabular}[c]{@{}l@{}}This statement contains factual information. However, \\ the general public would not be interested in checking \\ the presented factual claim.\end{tabular} \\ \midrule
But first of all, this is a nation of immigrants. & UFS & \begin{tabular}[c]{@{}l@{}}This statement contains factual information. However, \\ the general public would not be interested in checking \\ the presented factual claim.\end{tabular} \\ \midrule
\begin{tabular}[c]{@{}l@{}}I think everybody understands at this point that we are\\ experiencing the worst financial crisis since the Great\\ Depression.\end{tabular} & CFS & \begin{tabular}[c]{@{}l@{}}This statement contains both opinions and factual \\ information. The factual information is worthy of \\ veracity checking.\end{tabular} \\ \midrule
\begin{tabular}[c]{@{}l@{}}Government spending has gone completely out of  \\control; \$10 trillion dollar debt we're giving to our\\ kids, a half-a-trillion dollars we owe China.\end{tabular} & CFS & \begin{tabular}[c]{@{}l@{}}This statement is presenting data with a quantity. \\ People in general would be interested to know \\ whether the quantity is correct or not.\end{tabular} \\ \midrule
\begin{tabular}[c]{@{}l@{}}In the first place I've never suggested that Cuba was\\ lost except for the present.\end{tabular} & CFS & \begin{tabular}[c]{@{}l@{}}This statement is presenting a factual claim regarding \\ a past incident. People in general would be interested \\ to know whether the statement is true or false.\end{tabular} \\ \bottomrule
\end{tabular}
\caption{Training sample sentences along with their ground-truth labels, and explanations}
\label{tab:explanation}
\end{table*}

\subsection{Platform Development}
A rich and controlled data collection website~\footnote{\url{http://idir.uta.edu/classifyfact_survey}} was developed to collect the ground-truth labels of the sentences. Figure  \ref{fig:data_collection} shows its interface. A participant is presented one sentence at a time. The sentence is randomly selected from the set of sentences not seen by the participant before. The participant can assign one of three possible labels [NFS, UFS, CFS] for the sentence. If the participant is not confident to assign a label for a sentence, the sentence can be skipped. It is also possible to go back and modify previous responses. With just the text of a sentence itself, it is sometimes difficult
to determine its label. The interface has a ``more context'' button. When it is clicked, the system shows the four preceding sentences of the sentence in question which may help the participant understand
its context. We observe that, about 14\% of the time, participants chose to read the context before labeling a sentence.

\subsection{Recruitment Policy}
We recruited paid participants (mostly university students, professors and journalists who are aware of U.S. politics) using flyers, social media, and direct emails.

\begin{figure}[h!]
    \centering
    \includegraphics[width=0.99\linewidth]{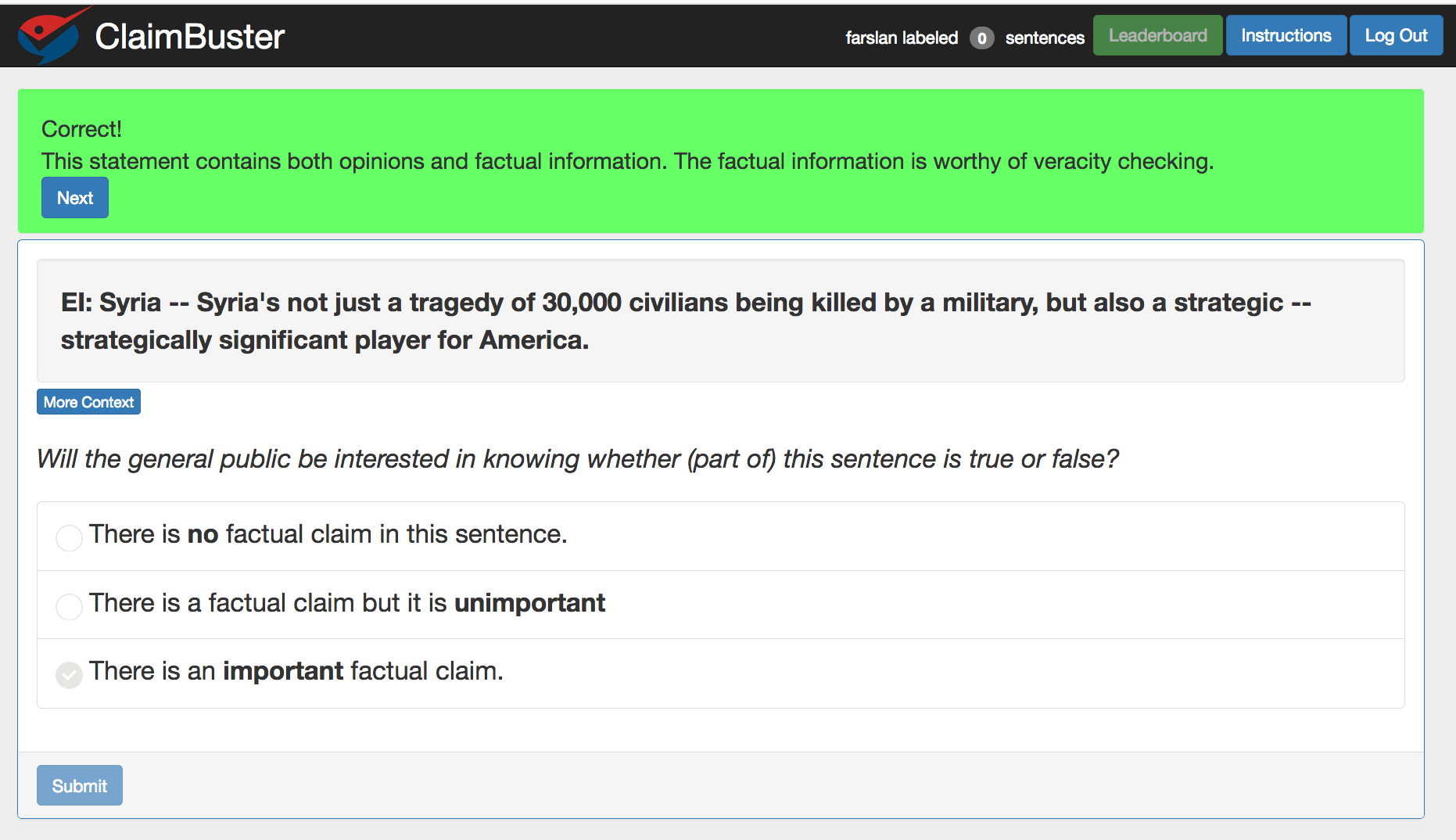}
    \caption{Participant training interface}
    \label{fig:participant_training}
\end{figure}

\begin{table*}[!ht]
\centering
\begin{tabular}{lcccc}
\toprule
 & Student & Professor & Journalist/Reporter &  Other \\
 \midrule
All Participants & 370 & 10 & 6 & 19 \\
Top-quality Participants & 86 & 5 & 4 & 6  \\
\bottomrule
\end{tabular}
\caption{\small Frequency distribution of participants' professions}
 \label{tab:participant_profession}
\end{table*}

\subsubsection{Participant Training}
We used $40$ labeled sentences to train all the participants. Each of these sentences was supplemented with an explanation regarding their labels from three experts.
Every participant must go through all these 40 sentences at the very beginning. After they label a sentence, the website will immediately disclose its ground-truth label and explain it (see Figure~\ref{fig:participant_training}). Some of the training sentences are shown in Table~\ref{tab:explanation}. Furthermore, we arranged multiple on-site training workshops for available participants. During each workshop, at least two experts were present to clear the doubts the participants may have about the data collection website and process. Through interviews with the participants, we observed
that these training measures were important in helping the participants achieve high work quality.

\subsection{Quality Control}
We selected 1032 sentences from all the sentences to create a ground-truth dataset. Three experts agreed upon the labels of these sentences: 731 NFS, 63 UFS, 238 CFS. We used this ground-truth dataset to detect spammers and low-quality participants for ensuring high-quality labels.
On average, one out of every ten sentences given to a participant (without letting the participant know) was randomly chosen to be a screening sentence. First, a random number decides the type (NFS, UFS, CFS) of the sentence. Then, the screening sentence is randomly picked from the pool of screening sentences of that particular type. The degree of agreement on screening sentences between a participant and the three experts is one of the factors in measuring the quality of the participant. For a screening sentence, when a participant's label matches the experts' label, s/he is rewarded with some points. If it does not match, s/he is penalized. We observe that not all kinds of mislabeling has equal significance. For example, labeling an NFS sentence as a CFS is a more critical mistake than labeling a UFS as a CFS. We defined weights for different types of mistakes and incorporated them into the quality measure. 

\begin{figure}[h!]
 \centering
\includegraphics[width=0.98\linewidth]{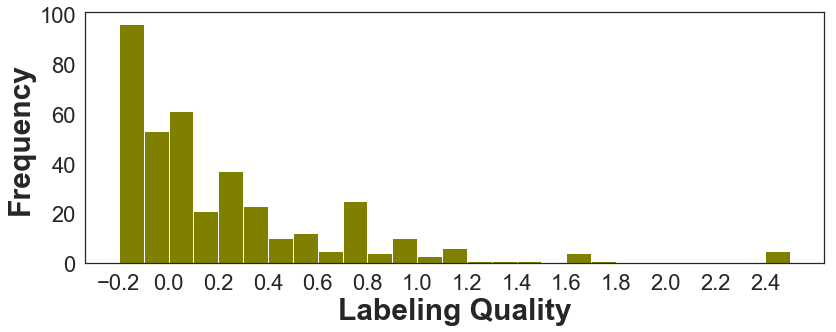}
 \caption{\small Frequency distribution of participants' labeling quality}
 \label{fig:participant_quality}
\end{figure}

\begin{figure}[!h]
\centering
  \begin{tikzpicture}
    \centering
    \begin{axis}[
        width  = 0.9*\linewidth,
        height  = 0.25*\textwidth,
        ylabel= {\textbf{Percentage}},
        enlargelimits=0.05,
        ybar=5*\pgflinewidth,
        x tick label style={font=\small, rotate=45},
        symbolic x coords={NFS\_UFS, NFS\_CFS, UFS\_NFS, UFS\_CFS, CFS\_NFS, CFS\_UFS },
        xtick align=inside,
        ymin=0,
        ymax=30,
        xtick=data]
    \addplot [fill=cfs]
        coordinates {(NFS\_UFS,21)(NFS\_CFS,10)
        (UFS\_NFS,17)(UFS\_CFS,25)(CFS\_NFS,13)(CFS\_UFS,14)};
    \end{axis}
    \end{tikzpicture}
  \caption{Error type distribution} 
  \label{fig:err_dist}
\end{figure}
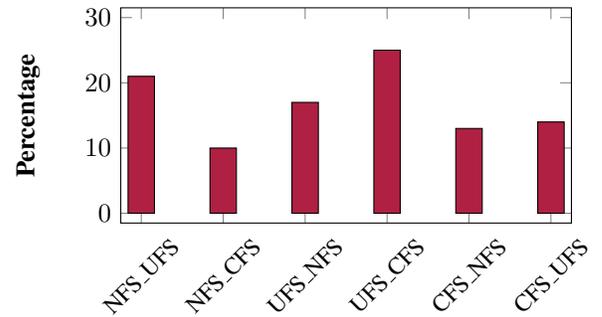

Formally, given $SS(p)$ as the set of screening sentences labeled by a participant $p$, the labeling quality of $p$ ($LQ_p$) is

 $$LQ_p = \frac{ \sum_{s \in SS(p)} \gamma^{lt} }{|SS(p)|}$$\\   
where $\gamma^{lt}$ is the weight factor when $p$ labeled the screening sentence $s$ as $l$ and the experts labeled it as $t$. Both $l,t \in \{NFS$, $UFS$, $CFS\}$. We set $\gamma^{lt} = -0.2$ where $l = t$, $\gamma^{lt} = 2.5$ where $(l,t) \in \{(NFS, CFS), (CFS, NFS)\}$ and $\gamma^{lt} = 0.7$ for all other combinations. The weights are set empirically. If $LQ_p \leq 0$ for a participant $p$ and $p$ labeled at least $50$ sentences, we designate $p$ as a top-quality participant. A total of $405$ participants contributed in the data collection process so far. Among them, $101$ are top-quality participants. Table~\ref{tab:participant_profession} depicts the distribution of all participants' and top-quality participants' professions where $91\%$ of all participants and $85\%$ of top-quality participants defined themselves as students. Figure~\ref{fig:participant_quality} shows the frequency distribution of $LQ_p$ for all participants. Throughout data collection process, the top-quality participants encountered screening sentences $9986$ times; $5222$ NFS, $1664$ UFS, and $3100$ CFS. They chose incorrect labels $511$ $(5\%)$ times. Figure~\ref{fig:err_dist} shows the percentages of six error types among these 511 cases. For instance, UFS\_CFS represents the cases in which participants mislabeled UFSs as CFSs. Besides, UFS\_CFS is the most frequent error type.

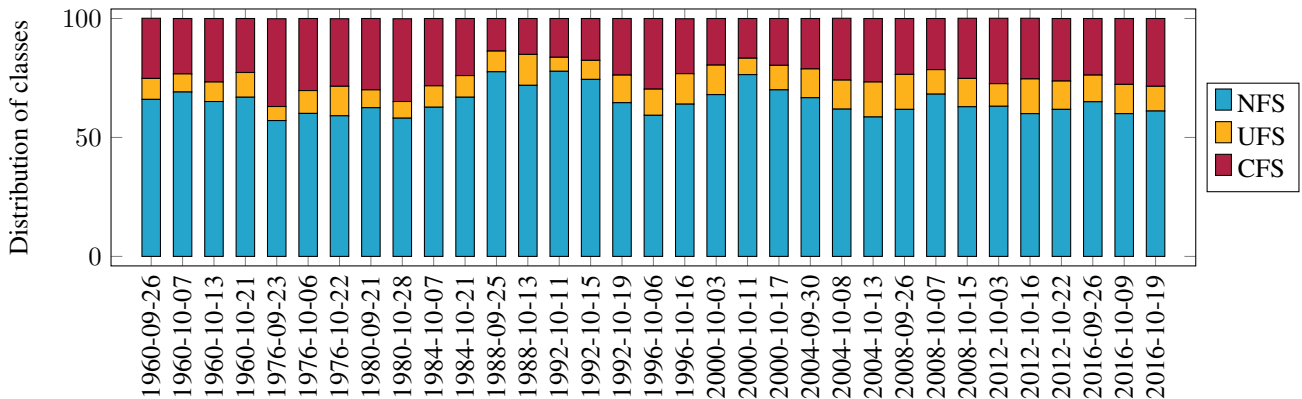
\begin{figure*}
    \centering
  \begin{tikzpicture}
  \begin{axis}[
    height=5cm,width=0.9\textwidth,
    ybar stacked,  
    bar width=7pt,
    enlargelimits=0.04,
    legend style={at={(1.01,0.5)},
      anchor=west},
    x tick label style={rotate=90},
    symbolic x coords={1960-09-26, 1960-10-07, 1960-10-13, 1960-10-21, 1976-09-23, 1976-10-06, 1976-10-22, 1980-09-21, 1980-10-28, 1984-10-07, 1984-10-21, 1988-09-25, 1988-10-13, 1992-10-11, 1992-10-15, 1992-10-19, 1996-10-06, 1996-10-16, 2000-10-03, 2000-10-11, 2000-10-17, 2004-09-30, 2004-10-08, 2004-10-13, 2008-09-26, 2008-10-07, 2008-10-15, 2012-10-03, 2012-10-16, 2012-10-22, 2016-09-26, 2016-10-09, 2016-10-19 }, 
    ymin=0,
    ymax=100,
    xtick=data,
    ylabel=Distribution of classes,
  ]
  \addplot [fill=nfs] coordinates {(1960-09-26,66.0 ) (1960-10-07,69.1) (1960-10-13,65.1) (1960-10-21,66.9) (1976-09-23,57.1) (1976-10-06,60.1) (1976-10-22,59.1) (1980-09-21,62.5) (1980-10-28,58.1) (1984-10-07,62.7) (1984-10-21,66.9) (1988-09-25,77.6) (1988-10-13,71.9) (1992-10-11,77.8) (1992-10-15,74.4) (1992-10-19,64.6) (1996-10-06,59.3) (1996-10-16,64) (2000-10-03,68) (2000-10-11,76.4) (2000-10-17,70) (2004-09-30,66.7) (2004-10-08,61.9) (2004-10-13,58.6) (2008-09-26,61.8) (2008-10-07,68.2) (2008-10-15,62.9) (2012-10-03,63.1) (2012-10-16,60) (2012-10-22,61.8) (2016-09-26,65) (2016-10-09,60) (2016-10-19,61.1)};
  \addplot [fill=ufs] coordinates {(1960-09-26, 8.8) (1960-10-07,7.6) (1960-10-13,8.2) (1960-10-21,10.4) (1976-09-23,5.9) (1976-10-06,9.6) (1976-10-22,12.4) (1980-09-21,7.5) (1980-10-28,7) (1984-10-07,9) (1984-10-21,9.1) (1988-09-25,8.7) (1988-10-13,13) (1992-10-11,5.9) (1992-10-15,8) (1992-10-19,11.6) (1996-10-06,11) (1996-10-16,12.8) (2000-10-03,12.4) (2000-10-11,6.9) (2000-10-17,10.3) (2004-09-30,12.1) (2004-10-08,12.2) (2004-10-13,14.7) (2008-09-26,14.7) (2008-10-07,10.3) (2008-10-15,11.9) (2012-10-03,9.5) (2012-10-16,14.6) (2012-10-22,11.9) (2016-09-26,11.2) (2016-10-09,12.3) (2016-10-19,10.4)};
  \addplot [fill=cfs] coordinates {(1960-09-26,25.3) (1960-10-07,23.3) (1960-10-13,26.7) (1960-10-21,22.7) (1976-09-23,36.9) (1976-10-06,30.3) (1976-10-22,28.4) (1980-09-21,30) (1980-10-28,34.8) (1984-10-07,28.3) (1984-10-21,24) (1988-09-25,13.7) (1988-10-13,15.1) (1992-10-11,16.3) (1992-10-15,17.6) (1992-10-19,23.8) (1996-10-06,29.7) (1996-10-16,23.1) (2000-10-03,19.6) (2000-10-11,16.7) (2000-10-17,19.7) (2004-09-30,21.2) (2004-10-08,26) (2004-10-13,26.7) (2008-09-26,23.5) (2008-10-07,21.5) (2008-10-15,25.3) (2012-10-03,27.5) (2012-10-16,25.5) (2012-10-22,26.3) (2016-09-26,23.8) (2016-10-09,27.7) (2016-10-19,28.5)};
  \legend{NFS,UFS,CFS}
  \end{axis}
  \end{tikzpicture}
 \caption{Class distribution per debate}  \label{fig:class_dist_per_debate}
\end{figure*}

\subsubsection{Incentives}
We devised a monetary reward program to encourage the participants to perform high-quality labeling. A participant $p$'s payment depends on their pay rate per sentence $R_p$ (in cents) and their number of labeled sentences. $R_p$ depends on $LQ_p$, the lengths of the labelled sentences, and the percentage of skipped sentences. The reason behind the later two factors is to discourage participants from skipping longer and more challenging sentences and to reward them for working on long, complex sentences. After multiple rounds of empirical analysis, we set $R_p$ as

$$R_p = \frac{L_p}{L}^{1.5}*(3-\frac{7*LQ_p}{0.2})*0.6^{\frac{|SKIP_p|}{|ANS_p|}}$$ \\

\noindent where, $L$ is the average length of all the sentences, $L_p$ is the average length of sentences labeled by $p$, $ANS_p$ is the set of sentences labeled by $p$ and $SKIP_p$ is the set of sentences skipped by $p$. The numerical values in the above equation were set in such a way that it would be possible for a top-quality participant to earn up to $10$ cents for each sentence.

The data-collection website also features a leaderboard which allows participants to see their rank positions by pay rate and total payment. This is designed to encourage serious participants to perform better and discourage spammers from further participation. Along with the leaderboard, the website provides helpful tips and messages from time to time to keep the participants motivated. 

\subsubsection{Stopping Condition}
A sentence $s$ will not be selected for further labeling if for $X \in \{NFS, UFS, CFS\}$, $\exists X$ such that $s_X \geq 2 \wedge s_X > (s_{NFS}+s_{UFS}+s_{CFS})/2$ where, $s_X$ denotes the number of top-quality labels of type $X$ assigned to $s$.

This condition ensures that a sentence has received a reasonable number of labels from top-quality participants and the majority of them agreed on a particular label. We assign the majority label as the ground-truth of that sentence.

\section{Dataset Description}
\label{sec-description}

\subsection{Dataset Statistics}
We collected $88,313$ labels among which $62,404$ $(70.6\%)$ are from top-quality participants. There are $22,281$ $(99.02\%)$ sentences which satisfy the above stopping condition. Table \ref{tab:class_distribution} shows the distribution of the classes in these sentences. The remaining 220 sentences, though, received many responses from top-quality participants, the labeling agreement did not satisfy the stopping condition. We assign each sentence the label with the majority count. Figure~\ref{fig:class_dist_per_debate} depicts the class distribution of sentences among 33 presidential debates, including all $22,501$ human-annotated sentences and $1,032$ expert labeled screening sentences.

\begin{table}[!h]
\centering
\begin{tabular}{lrr}
\toprule
Assigned label & \multicolumn{1}{c}{\#sent} & \multicolumn{1}{c}{\%} \\ \midrule
CFS & 5,318 & 23,87 \\ 
UFS & 2,328 & 10.45 \\ 
NFS & 14,635 & 65.68 \\ 
total & 22,281 & 100.00 \\ \bottomrule
\end{tabular}
\caption{Distribution of sentences over classes}
\label{tab:class_distribution}
\end{table}

During the data collection process, we advised the participants to skip the sentences that they are not confident in assigning a label. We analyzed the correlation between the number of sentences and the number of times they were skipped by the top-quality participants. We found that 17,874 (79.4\%) sentences were not skipped by any of the top participants, while the remaining 4,627 (20.6\%) sentences were skipped at least once. This observation indicates that participants found one in every five sentences challenging. Table~\ref{tab:user_skip} presents the distribution of these 4,627 sentences based on the frequency of them being skipped. For instance, $742$ sentences were skipped by any of the two top participants. One interesting observation is that the length of the sentences increased proportionally with the increasing number of skips. We examined the five sentences that were skipped most frequently, six and seven times, to probe whether the length of the sentence is what might account for this result. Table~\ref{tab:most_skipped} shows some descriptive information for these sentences. It can be observed that the last two sentences were labeled 14 and 15 times, respectively, to assign a label, although the last sentence contains only 19 words. This result indicates that the length of the sentence might not be the sole reason for the high number of skips of some sentences. The meaning of the sentence might play a significant role, too.

\begin{table}[!h]
\centering
\setlength\tabcolsep{3pt} 
\begin{tabular}{cccccccc} 
\toprule
\#skip & 1 & 2   & 3   & 4  & 5 & 6 & 7  \\ 
\midrule
\#sentence  & 3686 & 742 & 155 & 32 & 7 & 3 & 2  \\ 
\midrule
\#words(avg) & 19.3 & 21.3 & 25.2 & 25 & 26.7 & 35 & 62.5 \\
\bottomrule
\end{tabular}
\caption{Sentence distribution in terms of frequency of user skip }
\label{tab:user_skip}
\end{table}

\begin{table*}[!ht]
\centering
\setlength\tabcolsep{2.5pt} 
\begin{tabular}{ccccc} 
\toprule
\textbf{\#skip} & \textbf{sentence}  & \begin{tabular}[c]{@{}c@{}}\textbf{avg. length}\\\textbf{(in words)}\end{tabular} & \textbf{\#resp} & \begin{tabular}[c]{@{}c@{}}\textbf{assigned}\\\textbf{label}\end{tabular}  \\ \midrule
7  & \begin{tabular}[l]{@{}l@{}}You implement that NAFTA, the Mexican trade agreement, where they pay people \\a dollar an hour, have no health care, no retirement, no pollution controls, et cetera, \\et cetera, et cetera, and you're going to hear a giant sucking sound of jobs being \\pulled out of this country right at a time when we need the tax base to pay the debt \\and pay down the interest on the debt and get our house back in order.\end{tabular} & 77 & 5 & CFS \\
7 & \begin{tabular}[l]{@{}l@{}}Gene, there is a problem in the sense that there are some problem banks, and on De-\\cember 19th new regulations will go into effect which will in effect give the govern-\\ment the responsibility to close some banks that are not technically insolvent but \\that are plainly in trouble.\end{tabular}  & 48  & 2 & CFS \\
6 & \begin{tabular}[l]{@{}l@{}}We don't want to overreact, as the federal regulators have in my judgment, on good \\banks so that they've created credit crunches, that is, they have made our recession \\worse in the last couple of years -- but we do want to act prudently with the banks \\that are in trouble.\end{tabular}   & 51 & 3 & CFS  \\
6 & \begin{tabular}[l]{@{}l@{}}As a matter of fact, the statement that Senator Kennedy made was that - to the effect \\that there were trigger-happy Republicans, that my stand on Quemoy and Matsu was \\an indication of trigger-happy Republicans.\end{tabular} & 35 & 14  & CFS  \\
6 & \begin{tabular}[l]{@{}l@{}}I would like the record to show the panelists that Ross Perot took the first shot at the\\ press.\end{tabular}  & 19 & 15 & NFS  \\
\bottomrule
\end{tabular}
\caption{Descriptive information for the most skipped sentences}
\label{tab:most_skipped}
\end{table*}

We further analyzed each claim type by the number of top-quality participants' responses in labeling each sentence. Table~\ref{tab:response_distribution} depicts the distribution of responses over sentences --- the frequency of responses spans from 2 to 18. The vast majority of the sentences ($93\%$) were labeled by 2 or 3 participants. This means that at least two of the participants agreed upon the label. Four or five participants labeled $4.3\%$ of the remaining $7\%$ sentences. This indicates that at least three participants gave the same response. However, the participants were challenged to agree on the label of $620$ $(2.7\%)$ sentences as the number of the responses varies from 6 to 18.

\begin{table*}[!ht]
\centering
\begin{tabular}{ccccc} 
\toprule
\textbf{\#responses} & \textbf{\#sentences} & \textbf{NFS} & \textbf{UFS} & \textbf{CFS}  \\ 
\midrule
2  & 13057 (58\%)  & 9388 (63.9\%) & 845 (35.1\%) & 2824 (52.2\%)  \\
3   & 7865 (35\%) & 4545 (30.9\%) & 1192 (49.6\%) & 2128 (39.3\%)  \\
4  & 329 (1.5\%) & 224 (1.5\%) & 40 (1.7\%)  & 65  (1.2\%)     \\
5 & 630 (2.8\%) & 309 (2.1\%) & 152 (6.3\%)  & 169 (3.1\%)   \\
6-10 & 295 (1.3\%) & 125 (0.9\%)  & 70 (2.9\%) & 100 (1.8\%)   \\
11-18 & 325 (1.4\%) & 94 (0.6\%)  & 104 (4.3\%)  & 127 (2.3\%)    \\ 
\midrule
\textbf{Total} & 22501 & 14685 & 2403 & 5413          \\
\bottomrule
\end{tabular}
\caption{Frequency distribution of participants' responses over each class type }
\label{tab:response_distribution}
\end{table*}

\begin{figure}[!h]
\centering
\begin{tikzpicture}
\begin{axis}[
    ybar,
    enlargelimits=0.15,
    legend style={at={(0.5,-0.15)},
      anchor=north,legend columns=-1},
    ylabel={\#sentences},
    symbolic x coords={Democrat,Independent,Republican},
    xtick=data,
    xtick align=inside,
    nodes near coords,
    nodes near coords align={vertical},
    ]
\addplot [fill=nfs] coordinates {(Democrat,6221) (Independent,736) (Republican,7728)};
\addplot [fill=ufs] coordinates {(Democrat,1080) (Independent,91) (Republican,1232)};
\addplot [fill=cfs] coordinates {(Democrat,2650) (Independent,188) (Republican,2575)};
\legend{NFS, UFS, CFS}
\end{axis}
\end{tikzpicture}
  \caption{Distribution of claim types for parties of presidential candidates } 
  \label{fig:party_sent}
\end{figure}
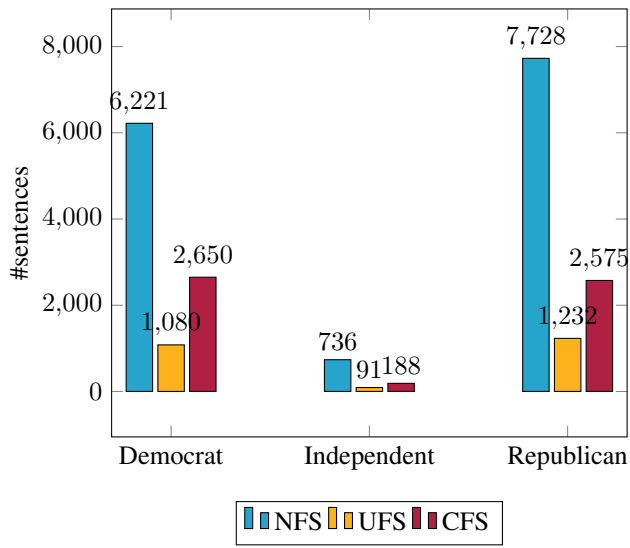

We also examined the relation between claim types and political parties. In 33 debate episodes, 69 presidential candidates took part, where 32 of them were from Democratic party, 33 were from Republican party,  while 4 were Independent candidates.
Figure~\ref{fig:party_sent} details the distribution of claim types made by each parties' presidential candidates. Republicans had the highest number of NFS while the Democrats had the highest number of CFS.

\subsection{Dataset Structure}
The ClaimBuster dataset consists of three files: \emph{groundtruth.csv} file, \emph{crowdsourced.csv} file, and \emph{all\_sentences.csv} file. The \emph{ground-truth} file contains only training and screening sentences whose labels were agreed upon by three experts. On the other hand, the \emph{crowdsourced} file consists of sentences that were labeled by top-quality participants. Both \emph{groundtruth} and \emph{crowdsourced} files are comprised of the same feature set, which is explained below. These two files can easily be merged to be leveraged in any study.

\noindent\textbf{Sentence\_id:} A unique numerical identifier to identify sentences in the dataset e.g., 1, 2, 31456.

\noindent\textbf{Text:} A sentence spoken by a debate participant. For example, ``Under my plan, I’ll be reducing taxes tremendously,from 35 percent to 15 percent for companies, small and big businesses.''

\noindent\textbf{Speaker:} Name of the person who verbalized the \emph{Text} (``Donald Trump'', ``Barack Obama'', ``Ronald Reagan'').

\noindent\textbf{Speaker\_title:} Speaker's job at the time of the debate e.g., ``Governor'', ``President'', ``Senator''.

\noindent\textbf{Speaker\_party:} Political affiliation of the \emph{Speaker} e.g., ''Democrat'', ``Republican'', ``Independent''.

\noindent\textbf{File\_id:} Debate transcript name e.g., ``2016-10-09.txt'', ``1960-09-26.txt''.

\noindent\textbf{Length:} Number of words in the \emph{Text}. For example, the length of the sample sentence provided for the ``Text'' feature is 20 words.

\noindent\textbf{Line\_number:} A numerical identifier to indicate the order of the \emph{Text} in the debate transcript e.g, 81, 82, 83.

\noindent\textbf{Sentiment:} Sentiment score of the \emph{Text}. The score ranges from -1 (most negative sentiment) to 1 (most positive sentiment). We used AlchemyAPI to calculate a sentiment score for
each sentence. We incorporated this feature into the dataset as it has been used as a feature in building machine learning models in these studies~\cite{hassan2017toward,hassan2017claimbuster}. This will enable researchers to compare their models with the models in the aforementioned studies.  

\noindent\textbf{Verdict:} Assigned class label (1 when the sentence is CFS, 0 when the sentence is UFS, and -1 when the sentence is NFS).

 \emph{All sentences} file contains all presidential debate sentences and not just the labeled ones. It has all the features shown above except for ``Verdict''. It also includes \emph{Speaker\_role} which depicts the role of the \emph{Speaker} in the debate as a participant e.g., Candidate, Moderator, Questioner. 

\section{Possible Use Cases}
\label{sec-usecase}
Recent years have witnessed a surge of interest in the scientific community to develop computational approaches to automate fact-checking components.  Some efforts have focused on assessing the truthfulness of a claim~\cite{ciampaglia2015computational,shi2016discriminative,leblay2017declarative,relevantdocumentdiscovery,evidencepatterns,huynh2019benchmark,exfakt}, although this work is in its infancy. A substantial number of studies~\cite{patwari2017tathya,hassan2017claimbuster,jimenez2018claimspot,konstantinovskiy2018towards}, on the other hand, have focused on detecting claims worthy of fact-checking from natural language statements. Early claim detection models rely on supervised classifiers such as SVM or logistic regression trained on hand-engineered features~\cite{hassan2017claimbuster,patwari2017tathya,claimrank}. Recent approaches, however, utilize neural network models~\cite{konstantinovskiy2018towards,jimenez2018claimspot,hansen2019neural,meng2020gradient}. A number of fact-checking organizations~\footnote{\url{https://fullfact.org/automated}}~\footnote{\url{https://team.inria.fr/cedar/contentcheck/}} \footnote{\url{https://reporterslab.org/tech-and-check/}} around the world make use of claim detection models in their fact-checking efforts to quickly detect claims to check. Claim detection is one particular task that can benefit from the \emph{ClaimBuster} dataset as some of the previous models~\cite{hassan2016comparing,hassan2017toward,hassan2017claimbuster,jimenez2018claimspot} used a subset of this dataset. The claim detection task can be approached in two ways. One of the approaches is to identify if a sentence comprises a factual claim aside from its check-worthiness. The second approach takes the check-worthiness of the claim into consideration. 
In the following sections, we argue how these two claim detection approaches can utilize the \emph{ClaimBuster} dataset.

\subsection{Factual Claim Detection}
This approach formulates the task as a binary classification task that identifies a sentence as either containing a factual claim (FC) or not containing a factual claim (NFC). This task can make use of the ClaimBuster dataset by combining UFS sentences and CFS sentences into FC sentences and using NFS sentences as NFC sentences. Then, a binary classifier can be trained on the FC and NFC sentences and applied to future sentences.

\subsection{Check-worthy Claim Detection} In order to prioritize the most check-worthy claims over less check-worthy ones, a check-worthiness score, which is the probability that a sentence belongs to the CFS class, is required. To this aim, this approach models the claim detection problem as a classification and ranking task. Given a sentence, a machine learning model or neural network model trained on the ClaimBuster dataset calculates a check-worthiness score that reflects the degree by which the sentence belongs to CFS.

\section{FAIRness}
\label{sec-fair}
In this section, we explain how we have made the ClaimBuster dataset adhere to the ``FAIR'' Facets: Findable, Accessible, Interoperable, and Re-usable.

To be \emph{Findable} and \emph{Accessible}, we make the dataset publicly available through Zenodo,~\footnote{\url{https://zenodo.org/}} a dataset sharing platform, allowing the complete dataset to be downloaded with the following citation. 

Fatma Arslan, Naeemul Hassan, Chengkai Li, \& Mark Tremayne. (2020). ClaimBuster: A Benchmark Dataset of Check-worthy Factual Claims [Dataset]. Zenodo. http://doi.org/10.5281/zenodo.3609356

The dataset files are provided in CSV (Comma Separated Values) format that can be utilized by any applications and exported to other data formats. The dataset is supplemented with a readme file explaining each data file in detail to optimize the re-use of the dataset.

\section{Conclusion}
\label{sec-conclusion}
In this paper, we present a dataset of claims from all U.S. general election presidential debates (1960 to 2016) along with the human-annotated check-worthiness label. We argue that the research community lacks a large labeled dataset of claims to leverage in claim detection tasks. To address this need, we provide a large dataset of $23,533$ sentences where each sentence is categorized into one of three categories; non-factual statement, unimportant factual statement, and check-worthy factual statement. One hundred one trained human-coders labeled these claims over a long period of two years. The ClaimBuster dataset is now publicly available to the research community.

\section{ Acknowledgments}
The work is partially supported by NSF grants IIS-1408928, IIP-1565699, IIS-1719054, OIA-1937143, a Knight Prototype Fund from the Knight Foundation, and subawards from Duke University as part of a grant to the Duke Tech \& Check Cooperative from the Knight Foundation and Facebook. Any opinions, findings, and conclusions or recommendations expressed in this publication are those of the authors and do not necessarily reflect the views of the funding agencies.


\bibliographystyle{aaai}

\end{document}